\documentclass[10pt,twocolumn,letterpaper]{article}
\usepackage[toc,page]{appendix}
\makeatletter
\@namedef{ver@everyshi.sty}{}
\makeatother
\usepackage{titling}
\usepackage{cvpr}              % To produce the CAMERA-READY version
\usepackage{graphicx}
\usepackage{amsmath}
\usepackage{amssymb}
\usepackage{float}
\usepackage{booktabs}
\usepackage{tablefootnote}
\usepackage{longtable}
\usepackage{tabularx}
\usepackage{multirow}
\usepackage{boldline}
\usepackage{verbatim}
\usepackage{graphicx}
\usepackage{pifont}
\usepackage[T1]{fontenc}
\usepackage{caption,subcaption,tikz}
\usepackage[pagebackref,breaklinks,colorlinks]{hyperref}
\usepackage[capitalize]{cleveref}
\crefname{section}{Sec.}{Secs.}
\Crefname{section}{Section}{Sections}
\Crefname{table}{Table}{Tables}
\crefname{table}{Tab.}{Tabs.}
\def\cvprPaperID{00002} % *** Enter the CVPR Paper ID here
\def\confName{V4AS@CVPR’22}
\def\confYear{2022}
\newcommand{\expect}[1]{\ensuremath{\operatorname{\mathbb{E}}\!\left[ #1 \right]}}
\newcommand{\var}[1]{\ensuremath{\operatorname{Var}\!\left[ #1 \right]}}

\newcommand{\minisection}[1]{\vspace{0.04in} \noindent {\bf #1}\ }

\begin{document}

\title{An Efficient Domain-Incremental Learning Approach to Drive in \\All Weather Conditions}

\author{M. Jehanzeb Mirza$^{1,2}$
\and
Marc Masana$^{1,3}$
\and
Horst Possegger$^1$
\and
Horst Bischof$^{1,2}$
\vspace{0.1cm}
\and
$^{1}$\text{Institute of Computer Graphics and Vision, Graz University of Technology.}\\ 
$^{2}$\text{Christian Doppler Laboratory for Embedded Machine Learning.}\\
$^{3}$\text{TU Graz - SAL Dependable Embedded Systems Lab, Silicon Austria Labs.}\\
{\tt\small $\{$muhammad.mirza, marc.masana, possegger, bischof$\}$@icg.tugraz.at}
}
\maketitle

%%%%%%%%% ABSTRACT
\begin{abstract}
Although deep neural networks enable impressive visual perception performance for autonomous driving, their robustness to varying weather conditions still requires attention. 
When adapting these models for changed environments, such as different weather conditions, they are prone to forgetting previously learned information. This \emph{catastrophic forgetting} is typically addressed via incremental learning approaches which usually re-train the model by either keeping a memory bank of training samples or keeping a copy of the entire model or model parameters for each scenario.
While these approaches show impressive results, they can be prone to scalability issues and their applicability for autonomous driving in all weather conditions has not been shown. 
In this paper we propose DISC -- \textbf{D}omain \textbf{I}ncremental through \textbf{S}tatistical \textbf{C}orrection -- a simple online \emph{zero-forgetting} approach which can incrementally learn new tasks (\ie~weather conditions) without requiring re-training or expensive memory banks. 
The only information we store for each task are the statistical parameters as we categorize each domain by the change in first and second order statistics.
Thus, as each task arrives, we simply `plug and play' the statistical vectors for the corresponding task into the model and it immediately starts to perform well on that task.
We show the efficacy of our approach by testing it for object detection in a challenging domain-incremental autonomous driving scenario where we encounter different adverse weather conditions, such as heavy rain, fog, and snow.
\end{abstract}

%%%%%%%%% BODY TEXT
\section{Introduction}
\label{sec:intro}
In order to entrust safety-critical systems such as autonomous vehicles with human lives, they must operate robustly in widely different environments.
While recent deep learning-based approaches achieve very strong performance, \eg~for object detection~\cite{redmon2018yolov3, ren2015faster, liu2016ssd,yolov4}, they are usually only trained and evaluated on data collected mostly in clear weather conditions~\cite{geiger2013vision,han2021soda10m,sun2020scalability}.
%Recent research advancements on deep learning-based approaches show very strong performance, \eg~for object detection~\cite{redmon2018yolov3, ren2015faster, liu2016ssd,yolov4}, when tested in data collected mostly in clear weather conditions~\cite{geiger2013vision,han2021soda10m,sun2020scalability}. 
However, a major real-world challenge for autonomous vehicles is to encounter deteriorating weather conditions while driving. As shown in~\cite{mirza2021robustness, michaelis2019benchmarking, bijelic2019recovering}, even slightly changing weather conditions can hamper the performance of object detectors considerably, making them unsafe for use in safety-critical systems. 

\begin{figure}
    \centering
    \includegraphics[width=.9\linewidth]{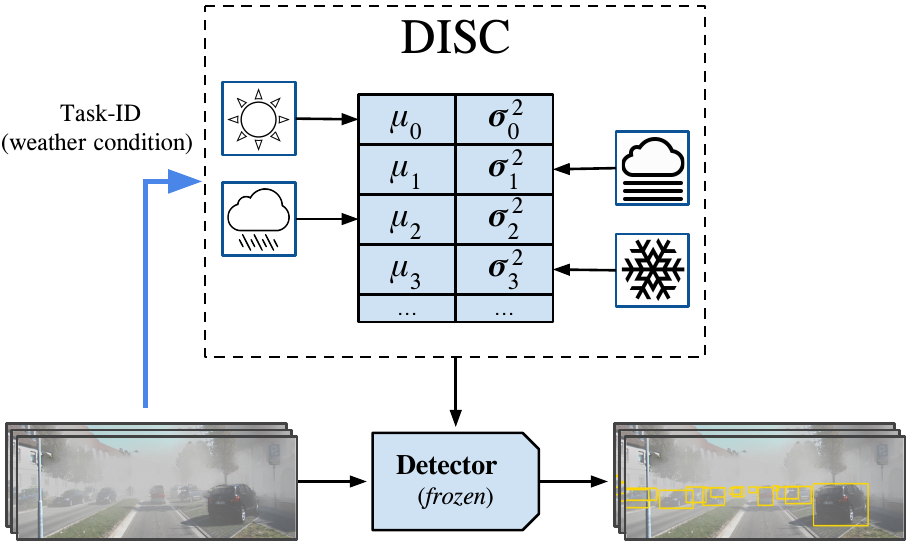}
    \caption{Robust object detection for autonomous driving across varying weather conditions with DISC. We only require domain-specific statistics (\ie~$\mu_{\text{ID}}, \sigma_{\text{ID}}^2$ per weather condition ID) which are inserted into a frozen detection model to achieve \emph{zero-forgetting}, \ie~the detection performance on previously encountered weather conditions does not degrade over time.}
    %OLD 01.04.:
    %\caption{Overview of DISC, our \emph{plug-and-play} approach, on a frozen YOLOv3 object detection architecture for domain-IL of different weather conditions. DISC only requires domain-specific statistics for different weather conditions in order to provide \emph{zero-forgetting} incremental learning.}
    \label{fig:motivation}
\end{figure}
A simple way to address such changing environmental conditions is to train different models for each individual weather condition.
However, this requires extensive data collection, careful manual annotation and time-consuming re-training.
Additionally, all these models need to be saved for future use which is sub-optimal because a significant amount of their learned weights might be redundant.

Instead of training a separate model for each different scenario, a better solution is to ideally have only a single model which can work in all weather scenarios.
%A better solution than training a separate model for each different scenario is to ideally have only a single model which can work in all weather scenarios. 
However, naively training a model for all weather scenarios individually can lead to a variety of problems, where the most critical is \emph{catastrophic forgetting}~\cite{french1999catastrophic, goodfellow2013empirical, kirkpatrick2017overcoming, mccloskey1989catastrophic}. This means that as deep neural networks are trained for new domains (\eg~weather conditions), they forget information from the previous domain. For example, if an object detector trained on clear weather conditions is re-trained with foggy weather data, its performance improves for foggy weather but degrades for clear weather.
This performance degradation during such model adaptation is typically addressed via Incremental Learning~(IL) approaches~\cite{vandeven2019three, delange2021continual, masana2020class}.
These approaches learn a \emph{sequence of tasks} (\eg~different weather conditions), one at a time, without having access to data from the previous tasks. They learn to perform good on a new task while retaining the performance on previously learned tasks. If performance does not degrade on previous tasks at all, then they are termed \emph{zero-forgetting} approaches.  

% Incremental learning approaches can be categorized into two distinct groups: class and domain incremental learning. Class-incremental learning approaches~\cite{masana2020class} consider different classes in a dataset as tasks. The goal is to incrementally learn new classes (tasks) without forgetting the previous classes (tasks). Domain incremental learning approaches~\cite{garg2022multi, nam2016learning, churamani2021domain}, on the other hand, consider different domains as tasks to learn in an incremental manner while the number of classes remains constant in these tasks.   
%
% An optimal solution to the underlying problems could be to design a system which retains information from all the domains it has been adapted or trained on, it is data efficient and also allows for dynamic adaptation if the need arises. However, to the best of our knowledge there is no work which tackles these problems together. 
%
% \revise{[Paragraph about previous work done (add some references from ITSC degrading weather paper), how do they do it, how inefficient are they, how difficult is it to apply them in autonomous driving scenario, why no system is available which solves incremental learning problem for autonomous driving?]}
%
% In the incremental learning setup solving the class incremental learning problem is perhaps the most common scenario explored by a lot of works~\revise{CITE}. In the class incremental learning setup the goal is to learn new classes in the 

Numerous approaches try to improve autonomous driving algorithms for degrading weather conditions. For example, foggy weather conditions can be handled by fusing multi-modal input data~\cite{bijelic2020seeing} or combining synthetic and real data~\cite{sakaridis2019guided, dai2018dark}. Similarly, several approaches address "driving in the wild", \eg~\cite{chen2021scale, chen2018domain_faster_rcnn, tran2019gotta}.
Although these approaches often provide state-of-the-art results for the individual task which they are specifically designed for, their applicability to different tasks and, in fact, to different weather scenarios is not yet clear. Furthermore, these approaches mostly follow the \emph{domain adaptation} paradigm which is different from the \emph{incremental learning} setup and are often prone to issues like catastrophic forgetting.
%OLD 01.04.:
%Numerous approaches exist which try to make autonomous driving algorithms robust to degrading weather conditions. For example, a multi-modal fusion approach is proposed in~\cite{bijelic2020seeing} for driving in foggy weather. In~\cite{sakaridis2019guided, dai2018dark} synthetic and real data is used for adapting to foggy weather conditions. Similarly, in~\cite{chen2021scale, chen2018domain_faster_rcnn, tran2019gotta} approaches for driving in the wild are proposed. Although these approaches often provide state-of-the-art results for the tasks which they are specifically designed for, their applicability to different tasks and in fact to different weather scenarios is not yet clear. Furthermore, these approaches mostly work in the domain adaptation paradigm which is different from the incremental learning setup and can often be prone to issues, such as catastrophic forgetting. 

We propose DISC, an efficient \textbf{D}omain-\textbf{I}ncremental learning approach which leverages \textbf{S}tatistical \textbf{C}orrection for robust object detection under varying weather conditions. DISC considers different weather conditions as distribution shifts and categorizes each condition according to its statistical difference.
We achieve zero-forgetting by only retaining the weather-specific first- and second-order statistics calculated by the detection model during training and replacing these statistics when we encounter a weather change, as shown in Figure~\ref{fig:motivation}.
%OLD: We show that by only retaining the first and second order statistics calculated by the network from different weather scenarios during training and only replacing these statistics in the network as a different weather scenario, for which we have prior information (Task-ID), we can achieve zero-forgetting. Our approach is shown in Figure~\ref{fig:motivation}.

% In this paper we propose DISC, an efficient domain incremental learning approach which considers different weather scenarios as distribution shifts. 
% In particular, we present a \textbf{D}omain-\textbf{I}ncremental learning approach leveraging \textbf{S}tatistical \textbf{C}orrection, named DISC, to robustly handle adverse weather conditions.
% We show that by only retaining the first and second order statistics calculated by the network from different weather scenarios during training and during testing only replacing these statistics in the network as a different weather scenario is encountered we can achieve zero-forgetting.

As we only store the domain-specific statistics of the model, our incremental learning approach is highly efficient, both in terms of computational cost and memory consumption, \ie~we neither need to store the entire model parameters nor any data samples from different domains.
%As we only store statistics of the model for each domain, our incremental learning approach is highly efficient, both in terms of computational cost and memory consumption compared to other approaches which store either model parameters \revise{[cite approaches]} or data samples \revise{[cite approaches]}. 
We show the applicability of our method in a challenging online setting where we let our system interact with multiple changes in domains (\ie~weather conditions) and show its effective zero-forgetting property. 

Our contributions can be summarized as follows: 
\begin{itemize}
    \item We propose an incremental learning approach which considers different weather conditions as distribution shifts and thus, only needs to store the first and second order statistics for each weather condition. 
    \item We show that by replacing the weather-specific statistics in an otherwise frozen model, we can efficiently realize a system which achieves zero-forgetting. 
    %\item We show that only by storing the statistics from different weather conditions and replacing them in the network when a certain weather condition arrives we can design a system which shows zero-forgetting.
    \item We demonstrate strong performance gains using our DISC in both offline and online learning scenarios, highlighting its effectiveness. 
\end{itemize}

\section{Related Work}
As our work lies at the intersection of incremental learning, unsupervised domain adaptation and autonomous driving in varying weather conditions, we summarize the current state-of-the-art in these research fields.
%In this section, we provide some background of the current state-of-the-art in these fields.

\subsection{Incremental Learning}
Incremental learning -- learning a sequence of tasks one after the other without access to previously learned tasks -- received increasing interest in recent years. Incremental learning can be divided into three main scenarios~\cite{hsu2018re, vandeven2019three}: task-incremental learning~\cite{delange2021continual}, class-incremental learning~\cite{mai2022online, masana2020class} and domain-incremental learning~\cite{acharya2020rodeo}. Throughout all these scenarios, existing approaches can be assigned to three main categories, namely replay-based, regularization-based and parameter isolation approaches.

Replay-based approaches explicitly store samples from previous tasks in a limited exemplar memory that can be rehearsed during the training of new tasks~\cite{rebuffi2017icarl, rolnick2019experience, isele2018selective, chaudhry2019continual}. Alternatively, some methods substitute the replay based on an exemplar memory by learning a generative model which is capable of describing previous distributions and sample from them~\cite{shin2017continual, atkinson1802pseudo, lavda2018continual, ramapuram2020lifelong}. To counter the potential overfitting to the replayed memories, some approaches propose to constrain the interference between the new task and all previously learned ones~\cite{chaudhry2018efficient, aljundi2019gradient, lopez2017gradient}.

Regularization-based approaches introduce extra regularization terms in the training loss which limit the change of important weights or large shifts on the activations. One of the most common strategies is knowledge distillation, which enforces the outputs of the network to shift as little as possible for previously learned classes, while allowing enough change to learn the new ones~\cite{silver2002task, li2017learning, jung2016less, zhang2020class, rannen2017encoder}. The second most common approach consists of calculating the importance of each parameter in the network and penalizing their update based on that importance~\cite{kirkpatrick2017overcoming, liu2018rotate, aljundi2018memory, zenke2017continual}.

Parameter isolation approaches mostly assume no constraints on the model size. They usually isolate or freeze important model parameters from previous tasks and allow the models to introduce new parameters in order to exploit strong connections and avoid forgetting~\cite{aljundi2017expert, rusu2016progressive, xu2018reinforced, rajasegaran2019random}. Other approaches enforce zero-forgetting by learning masks or paths for each parameter or each layer representation~\cite{mallya2018packnet, fernando2017pathnet, serra2018overcoming, masana2021ternary}.

% Maybe introduce GDumb and how it challenged some of the advances on those scenarios??? And how we also propose a simple method for solving the domain-IL scenario.
Most of the incremental learning approaches are usually proposed for either task-IL or class-IL, and in general are framed inside a classification problem. % WIP
In this paper, however, we demonstrate how to leverage domain-IL for adaptation to varying weather conditions.
% Final conclusion for IL related work:
To this end, we propose DISC, a method which is designed for domain-IL and which does not rely on replay memories, regularization terms, or even long training sessions. 
%In this paper we propose DISC, a method which is designed for domain-IL and which does not rely on replay memories,  regularization terms, or even long training sessions. 
Contrary to replay- or regularization-based methods, it provides zero-forgetting properties, and unlike parameter isolation models, it does not require the computationally expensive training or calculation of masks and paths.

\begin{figure*}
    \centering
    \includegraphics[scale=0.85, trim=8 4 5 3, clip]{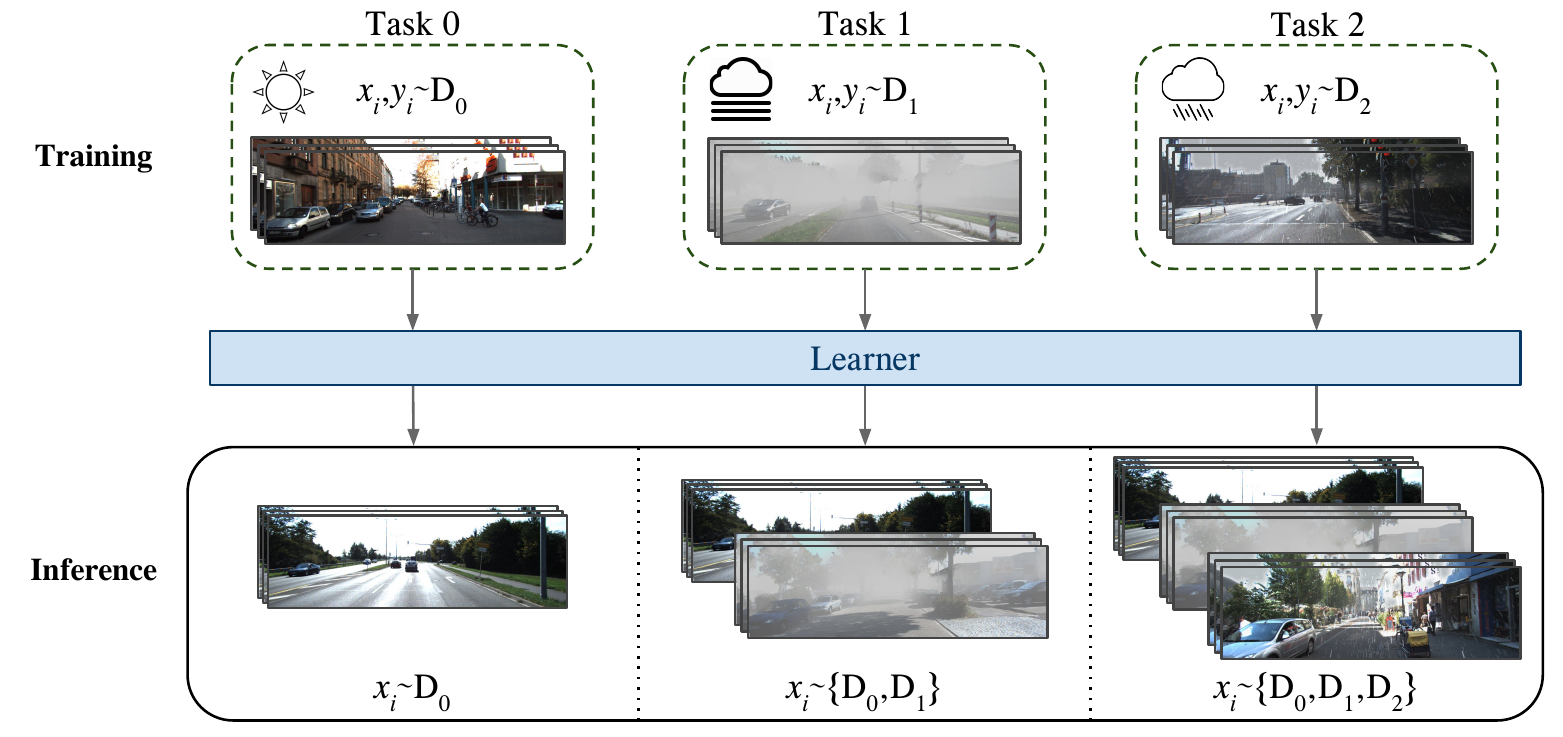}
    \caption{Domain-incremental learning setting used in this paper. Different tasks correspond to different weather conditions which are learned in the training phase.
    During inference, we evaluate the performance of the current task along with all the previously seen tasks.
    %During inference the performance of the current task along with all the previous tasks seen is evaluated. 
    }
    \label{fig:experimental_setup}
\end{figure*}

\subsection{Correcting Domain Statistics} 
Adapting the batch normalization~\cite{ioffe2015batch} statistics for the target domain data has been extensively used for unsupervised domain adaptation. For example, Li \etal~\cite{li2016revisiting} recalculate the first and second order statistics of the batch normalization layer for domain adaptation, while Carlucci \etal~\cite{maria2017autodial} learn hyper-parameters during training to optimally mix the statistics from source and target domain. Other approaches, such as~\cite{singh2019evalnorm, nado2020evaluating, schneider2020improving, arm} propose prediction-time batch normalization in order to reduce the statistical discrepancies between source and target domains. In addition to correcting the first and second order statistics for the target domain, Wang \etal~\cite{wang2020tent} also learn the scale and shift parameters of the batch normalization layer by calculating the gradients from the entropy of predictions. Mirza \etal~\cite{mirza2021norm} introduce DUA, a highly data-efficient method to adapt the statistics in an online manner for the target domain for unsupervised domain adaptation.

\subsection{Autonomous Driving in Varying Weather} 
Improving the robustness of autonomous vehicles to varying weather conditions also gained increased attention in recent years. Recent works propose approaches for semantic segmentation~\cite{sakaridis2019guided, dai2018dark} and image defogging~\cite{yan2020nighttime} during night. Sakaridis \etal~\cite{sakaridis2018semantic} improve semantic segmentation for foggy scenes using synthetic data. Bijelic \etal~\cite{bijelic2020seeing} propose a multi-modal fusion architecture for driving in foggy conditions. Chen \etal~\cite{chen2021scale, chen2018domain_faster_rcnn} propose object detection approaches for driving in the wild. RoyChowdhury \etal~\cite{roychowdhury2019automatic} propose a self-training approach for adaptation of object detectors to different weather conditions. Other works, such as~\cite{Shen_2020_CVPR_Workshops,kim2019bidirectional} propose image de-hazing approaches in the context of autonomous driving. 

% Some other works in this domain include~\cite{roychowdhury2019automatic} which is tested for object detection in degrading weather, for image de-hazing~\cite{Shen_2020_CVPR_Workshops,kim2019bidirectional} and for benchmarking robustness in degrading weather~\cite{bijelic2019recovering, mirza2021robustness, michaelis2019benchmarking}.

% ~\cite{ roychowdhury2019automatic, kim2019bidirectional, haurum2019raining, tran2019gotta, khan2019procsy, Shen_2020_CVPR_Workshops, Hirohashi_2020_CVPR_Workshops}.

Our work also focuses on driving in diverse weather conditions. 
However, in contrast to other task- and weather-specific approaches, \eg~\cite{sakaridis2019guided, dai2018dark, bijelic2020seeing, chen2021scale, chen2018domain_faster_rcnn}, which mostly follow the \emph{domain adaptation} paradigm, we propose a \emph{zero-forgetting domain-incremental learning} approach which learns to drive in varying weather conditions as they are encountered while not reducing the performance on previous conditions.
We consider different weather conditions as distribution shifts and employ DUA~\cite{mirza2021norm} to learn the weather-specific statistics. 
During testing we only need to replace these weather-specific statistics which makes our approach highly efficient and achieve zero-forgetting. 
%During testing we `plug in' these weather-specific statistics. As we do not change any other parameters in our network and only replace the statistics thus our methodology is very efficient and achieves zero-forgetting.  
%
% Our work also focuses on driving in adverse weather conditions, however, in contrast to other approaches, such as~\cite{sakaridis2019guided, dai2018dark, bijelic2020seeing, chen2021scale, chen2018domain_faster_rcnn} which are task and weather specific and their applicability in different weather conditions remains a question, we propose a \textit{zero-forgetting} domain incremental learning approach which learns to drive in all weather conditions.

% ----------------------------------------------

\section{Domain-IL for Weather Conditions}
\label{sec:domain-il}
In incremental learning scenarios, a sequence of tasks is learned one at a time within their own training sessions, without access to data from previously seen tasks. Following most common incremental scenarios~\cite{vandeven2019three, delange2021continual, masana2020class}, we can define an IL problem as the sequence of $n$ tasks:
\begin{equation}
    \mathcal{T} \ =\ [(C^{1}, D^{1}), (C^{2}, D^{2}), \ \ldots\ , (C^{n}, D^{n})],
\end{equation}
where the set of classes $C^t\!=\!\{c^t_1,c^t_2...,c^t_{n^t}\}$ represents each task $t$, learned with training data $D^{t}$. The training data consists of input features $x\in X$ (\ie~images in our setup) and corresponding ground truth labels $y$ (\ie~bounding box annotations and class labels in our case). Therefore, we have $D^{t}\!=\!\{(x_{1}, y_{1}),(x_{2}, y_{2}),\ \ldots\ ,(x_{k}, y_{k})\}$ all available pairs of input and label for a given task, where all $y_{i}\in C^{t}$.

In this paper, we focus on the variant of \emph{domain-incremental} learning, where all tasks share the same classes, \ie~\mbox{$C^{t}\!=\!C^{1} \ \forall  t\!\in\!\mathcal{T}$}. As the sequence of tasks is learned, the same objects have to be detected while their domain and data distribution changes. At inference time, similar to task-incremental learning~\cite{delange2021continual}, we have access to the task-ID. For our use-case this means that we know the current weather condition. 

The properties from the proposed domain-incremental learning scenario correspond to the real-world problem of object detection under different weather conditions. We aim to learn the same set of objects (classes) over different weather conditions (domains), while having access to the task-ID (current weather). A representation of the domain-IL scenario we adopt is depicted in Figure~\ref{fig:experimental_setup}.

% ----------------------------------------------
\section{Method}
\label{sec:method}
Before presenting our domain-IL approach DISC, we first explore how correcting task-specific latent representation statistics can be used for domain adaptation.
%OLD - In this section, we first explore how correcting task-specific latent representation statistics can be used for domain adaptation. After, we describe our proposed domain-IL approach.
% OLDer - We first summarize how correcting the domain statistics can be used for domain adaptation in Section~\ref{subsec:correcting_bn}. Section~\ref{subsec:domain_incremental_learning} describes our domain incremental learning setup.

\subsection{Statistical Correction for Domain Adaptation}
\label{subsec:correcting_bn}
In deep neural networks, batch normalization~\cite{ioffe2015batch} is a strategy that aims to reduce the internal covariate shifts between each layer's input distribution during training. Essentially, it normalizes the input activations of each layer to have a zero-centered mean and unit variance, such that
% OLD - In deep neural networks, batch normalization~\cite{ioffe2015batch} was proposed to reduce the internal covariate shifts between the distributions of inputs to each layer during training. Essentially, it normalizes the input activations of each layer to have a zero-centered mean and unit variance via
\begin{equation}
 \hat{x} = \frac{x - {\expect{{X}}}}{\sqrt{{\var{{X}}} + \epsilon}} \cdot \gamma + \beta,
  \label{eq:bn}
\end{equation}
where each input sample $x$ is normalized by the mean and variance calculated from the activations during training.
%coming from the training data \mbox{$X\!=\!\{x_{1},\ \ldots\ ,x_{k}\}$} with \mbox{$x_{i}\!\sim\!D$}. 
The parameters $\gamma$ and $\beta$ denote the distribution scale and shift, respectively, while \mbox{$\epsilon>0$} prevents division by zero.
% OLD - where each input sample $x$ is normalized by the mean and variance calculated from the activations coming from the training data $X$. The $\gamma$ and $\beta$ are the scale and shift parameters, respectively, while $\epsilon>0$ prevents division by zero. 
%The expected value {\expect{{X}}} and \var{{X}} of the training statistics are calculated by a running estimate over the entire training data coming in batches,
The expected value {\expect{{X}}} is estimated by the running mean
\begin{equation}
  {\hat\mu_{k}} = (1-\rho)\cdot{\hat\mu_{k-1}} + \rho \cdot {\mu_{k}},
  \label{eq:running_mean}
\end{equation}
while the variance \var{{X}}, \ie~the variability around the expectation is estimated by the running variance
\begin{equation}
  {\hat\sigma_{k}^2} = (1-\rho)\cdot{\hat\sigma_{k-1}^2} + \rho \cdot \sigma_{k}^2,
  \label{eq:running_var}
\end{equation}
where $\mu$ and $\sigma^2$ denote the mean and variance of the current incoming batch of data samples, respectively. The running mean and variance are updated with each new batch of data. The momentum parameter $\rho$ (default $\rho=0.1$) controls the balance between the previously accumulated statistics and the effect of the current batch statistics.

By design, the behavior of batch normalization is inherently different between training and inference phases. During training, since data is forwarded through the network in batches, the statistics are updated by the batch normalization layer via the running mean and variance (Eqs.~\eqref{eq:running_mean} and~\eqref{eq:running_var}).
% OLD - By design, the behavior of batch normalization is inherently different for training and testing. During training, as data can be obtained in batches, the statistics are calculated and updated by the batch normalization layer via the running mean and variance (Eqs.~\ref{eq:running_mean} and~\ref{eq:running_var}).
However, at inference time, the statistics accumulated during training are fixed and used for normalization following Eq.~\eqref{eq:bn}. Batch normalization works best when both train and test data are from a similar distribution~\cite{he2016deep, he2019multi, xie2017aggregated}. However, when train and test data distributions are not well aligned, batch normalization can cause performance degradation~\cite{wu2021rethinking}.
% OLD - During evaluation, the statistics calculated during training are fixed and used for normalization. Batch normalization works well in practice when training and testing data are in-distribution~\cite{he2016deep, he2019multi, xie2017aggregated}. However, when training and testing data is out-of-distribution, batch normalization can cause performance degradation~\cite{wu2021rethinking}.
% One way to overcome this issue is to correct the domain statistics and directly adapt these statistics on the out-of-distribution test data which has been explored by~\cite{li2016revisiting,  wang2020tent, schneider2020improving, nado2020evaluating, arm, mirza2021norm}. Essentially, these approaches either adapt or replace the estimated statistics by the batch normalization layer during training for the out-of-domain test data and show considerable improvements. In our work we use~\cite{mirza2021norm} due to its data efficient nature for adaptation. 

Correcting the statistics when train and test data distributions are not aligned can be helpful for unsupervised domain adaptation~\cite{li2016revisiting,  wang2020tent, schneider2020improving, nado2020evaluating, arm, mirza2021norm}. Existing approaches either adapt or replace the statistics estimated by the batch normalization layers during training for the out-of-distribution test data, improving performance on the latter. Since different weather conditions can also be considered as being out-of-distribution \wrt~the clear weather condition, we propose a similar technique for our domain-incremental learning through statistical correction approach. In particular, we draw inspiration from DUA~\cite{mirza2021norm} due to its data-efficient nature for domain adaptation.
% OLD - As shown in~\cite{li2016revisiting,  wang2020tent, schneider2020improving, nado2020evaluating, arm, mirza2021norm}, correcting the domain statistics for the out-of-distribution test data can be helpful for unsupervised domain adaptation. Essentially, these approaches either adapt or replace the statistics estimated by the batch normalization layer during training for the out-of-distribution test data and show considerable improvements. Since adverse weather conditions can also be considered out-of-distribution \wrt~the clear weather condition, we take inspiration from these approaches. In particular, we use DUA~\cite{mirza2021norm} due to its data efficient nature for adaptation. 

% The effect of batch normalization has been extensively studied in \cite{wu2021rethinking, santurkar2018does} and it has been shown that in addition to introducing stability to the training process it also helps to improve the performance on the downstream tasks~\cite{he2016deep,xie2017aggregated}. 
\subsection{DISC}
\label{subsec:domain_incremental_learning}
Motivated by the promising results for domain adaptation, we leverage task-specific statistical correction for domain-incremental learning.
Specifically, we propose that each time a new task from the sequence arrives, we freeze the parameters of the model and only calculate the running mean and variance at each batch normalization layer.
While DUA~\cite{mirza2021norm} calculates the statistics from a fraction of the data for domain adaptation, for domain-IL we need to store task-specific statistics to overcome catastrophic forgetting.
%Similar to how DUA~\cite{mirza2021norm} calculates the statistics from a fraction of the data for domain adaptation, we adopt this strategy for domain-IL. However, in our case we need to store the statistics for each of the tasks.
%OLD - Motivated by the previous section, we aim to estimate the statistics for each task of the domain-IL setting. We propose that each time that a new task from the sequence arrives, we freeze the parameters of the model and only calculate the running mean and variance at each layer representation. Similar to how DUA~\cite{mirza2021norm} stores statistics from a fraction of the data at each layer for domain adaptation, we adopt the strategy for domain-IL. However, in our case we need to store those statistics for each of the tasks.
% OLDer - For our proposed domain-incremental learning approach, we assume that we are given a model $\Phi\clear$ trained solely on the clear weather data $X\clear\sim D^{0}$. In our experiments, we refer to this as the \emph{clear} condition. Further, we are also given a fraction of unlabeled data from degrading weather conditions $X\degraded$, \eg~fog, rain and snow. Our goal is then to change the model upon weather condition shifts such that it starts to perform better in the corresponding adverse weather scenarios.

Since our DISC only requires a quick, efficient, zero-forgetting storage of the statistics at each batch normalization layer for each task in the sequence, we divide its usage into two phases, namely adaptation \& plug-and-play:
% ---
% We classify the different weather conditions encountered in the environment as distribution shifts. Thus, we use DUA~\cite{mirza2021norm} in order to adapt the first and second order statistics of each distribution (\ie weather condition) which are estimated by the batch normalization layers. Our method has two distinct phases: 
% OLD - Changing weather conditions lead to distribution shifts in the input images. 
% OLD - While some characteristics stay the same, such as the environmental layout (\eg~lanes, buildings, sidewalks, etc.), other characteristics of the scene change according to the weather condition, \eg~illumination, contrast or visual appearance.
%The environmental layout stays the same (\eg lanes, buildings), but certain characteristics of the scene, such as illumination, contrast or appearance changes. 
% OLD - Even though this distribution shift may not seem severe, it is still sufficient to cause a drastic degradation of object detection performance as shown in~\cite{michaelis2019benchmarking, mirza2021robustness}. Since the domain gap between clear and adverse weather conditions is not huge and we need fast adaptation to changing weather, we leverage DUA~\cite{mirza2021norm} for the efficient statistical correction. In particular, our method can be divided in two distinct phases:
% ---

%\minisection{Adaptation phase:}
In the \textbf{Adaptation phase}, we require only a tiny fraction of unlabeled data from each weather condition to adapt the statistics estimated by the batch normalization layers via DUA~\cite{mirza2021norm}. DUA adapts the statistics by directly using the out-of-distribution test data without requiring back-propagation. It uses an adaptive momentum ($\rho$) scheme in order to adjust the statistics by using less than~1\% of test data. After adapting to each of the weather conditions until convergence (\ie~when the statistics become stable), we store the weather-specific statistics calculated by this model and discard the remaining model parameters. 
%We only need to store the vectors of the running mean $\hat\mu$ and variance $\hat\sigma^2$ from each batch normalization layer in the network. Unlike DUA, we do not use test data to adapt the statistics, we only use a fraction of the train data collected in the current task for adaptation.
This means that we only need to store the vectors of the running mean $\hat\mu$ and variance $\hat\sigma^2$ for each batch normalization layer in the model. In contrast to DUA, we do not use the test data to adapt the statistics, but only use the training data collected in degrading weather to calculate the statistics.
% OLD - \subsection*{Adaptation Phase}
% OLD - In this phase we leverage a tiny fraction of unlabeled data~$X\degraded$ to adapt the statistics calculated by the batch normalization layers via leveraging DUA. DUA adapts the batch normalization statistics by directly using the out-of-distribution test data without requiring back-propagation. It uses an adaptive momentum ($\rho$) scheme in order to adapt the statistics by using less than~1\% of test data. After adapting to each of the weather conditions until convergence we store the statistics calculated by the adapted model and discard the other parameters of the model. This is essentially equivalent to storing only the vectors of the running mean $\hat\mu$ and variance $\hat\sigma^2$ from each batch normalization layer in the network. Contrary to DUA setup, we do not use the testing data to adapt the statistics. In fact, we only use the training data collected in degrading weather for adaptation.

%\minisection{Plug-and-play phase:}
In the \textbf{Plug-and-Play phase}, we apply the weather-specific statistics vectors to the initial model, which is only trained on the first task. At inference, we replace the batch normalization statistics in the initial model with the ones learned for the current task (\emph{plug}) and forward the data through the network (\emph{play}). We do not replace or modify any other model parameters other than the running mean and variance of the batch normalization layers. Thus, we can exactly recover the model's original performance on all previous tasks, even after encountering new tasks. Consequently, DISC belongs to the family of \emph{zero-forgetting} incremental learning approaches~\cite{fernando2017pathnet, masana2021ternary, mallya2018packnet, mallya2018piggyback}.

The proposed approach provides several advantages. First, the lack of update on the model parameters completely avoids catastrophic forgetting when learning new tasks, effectively turning our proposed approach into a zero-forgetting one. Second, only having to calculate the statistics means that we do not need computationally expensive training sessions. In particular, only a single forward pass through the model with a fraction of the current unlabeled training data is sufficient. This makes our approach extremely efficient and easily adaptable. Third, the storage of the statistics for each layer grows linearly with the number of tasks, but it is still quadratically more efficient than storing parameters for each weight or a memory of exemplars, as most IL methods do. Next, since the estimate of the statistics is only based on the available samples from $D^{t}$, our method can be compared under both online and offline domain-IL scenarios. Finally, our approach only relies on the $x_{i}$ samples from the data, without the need for the labels $y_{i}$ -- except for the starting task model of the sequence which might require supervised training. This allows our method to be applicable in practical scenarios where the subsequent tasks in the sequence (\ie after the first), can be learned in an unsupervised manner.
% OLd - For this, we only need access to the tiny fraction of unlabeled data collected in these weather conditions. This makes our approach suitable for real-world autonomous driving scenarios, because we do not require carefully manually labelled samples for adaptation. Unlike the classical domain adaptation paradigm we leverage incremental learning. Adjusting to weather changes using domain adaptation approaches is prone to suffer from catastrophic forgetting and thus, after adaptation the model will perform worse on the previously learned weather scenarios. To overcome this issue, our goal is to incrementally learn to improve the model performance on the new weather condition without forgetting the previous weather condition.

A requirement of DISC and the proposed domain-IL scenario is to have access to the task-ID during inference, as it is done in task-incremental learning approaches~\cite{delange2021continual}. However, especially for weather conditions, this task-ID can easily be inferred by either training a separate classifier, or simply querying additional sensors available in modern cars, such as (typically infrared-based) rain sensors.
We also provide results for using incorrect task-IDs in Sec.~\ref{subsec:forward_backward_transfer}.

\begin{table}
\setlength\tabcolsep{2.7pt}
\small
  \centering
  \begin{tabular}{c|cccccccc|c}
  \toprule
  &car&van&truck&ped&sit&cyc&tram&misc&sum\\
  \midrule
  train&12884 &1307 &475 &1905 &101 &688 &258&398 &18016\\
  val& \phantom{0}1381&\phantom{0}150 &\phantom{0}54 &\phantom{0}262 &\phantom{0}16 &\phantom{0}83 & \phantom{0}34&\phantom{0}67&\phantom{0}2047\\
  test&14477 &1457 &565 &2320 &105 &856 &219&508&20507\\
  \bottomrule
  \end{tabular}
  \caption{Annotated instances of classes for the different dataset splits of KITTI used in our experiments.}
  \label{tab:dataset_statistics}
\end{table}

\section{Experimental Setting}
In this section, we first introduce the dataset and tasks on which we evaluate our domain-incremental learning approach. Then, we provide details about the experimental protocols and the baselines we compare to.

\subsection{Dataset and Tasks}
\label{subsec:dataset-tasks}
We conduct all our experiments on the widely used KITTI~\cite{geiger2013vision} autonomous driving dataset. The KITTI dataset consists of 8 object classes manually annotated for training both 2D and 3D object detectors. The dataset consists of real-world driving scenes captured in Germany in both rural and urban areas. Separate training and testing splits are provided for the KITTI dataset. However, publicly available annotations are only provided for the training split, whereas the testing split is used for the KITTI object detection challenge which is evaluated through their private server.

Thus, we follow the common protocol~\cite{ku2018joint, lang2019pointpillars} and divide the 7,481 images in the KITTI training dataset into train (3,740 images) and test (3,741 images) splits. Further, we use 10\% of this train split as validation split during training to optimize hyper-parameters. Details about the exact number of instances present for each class in each split is provided in Table~\ref{tab:dataset_statistics}. These splits are fixed for both training and inference across all experiments and methods. 

Recently, several works introduced approaches to change the input images realistically to simulate driving in different weather scenarios, such as fog~\cite{halder2019physics}, rain~\cite{tremblay2020rain}, and snow~\cite{hendrycks2019robustness}. 
We use these approaches to create different weather scenarios for our domain-incremental learning setup. Thus, we define the following four-task scenario:

\minisection{Task 0 - Clear:} 
The initial task in our incremental learning setup is driving in clear weather. Since the KITTI dataset is recorded almost exclusively in bright daylight conditions, we use the original KITTI data as input for this task.

\minisection{Task 1 - Fog:} The second task in the sequence is driving in foggy weather. Halder \etal~\cite{halder2019physics} introduce realistic physics-based rendering of fog. In particular, they use the depth information from the LiDAR sensor, re-projected into the camera view to simulate 7 different fog levels. 
The fog severity is categorized by the visibility in meters and varies from $30\,\text{m}$, corresponding to the most severe fog, to a visibility of $750\,\text{m}$, corresponding to the least severe fog. For this task, we apply their method to simulate fog on KITTI at the maximum severity level, \ie~$30\,\text{m}$ visibility.

\minisection{Task 2 - Rain:} The third task to learn is the rainy weather condition. Tremblay \etal~\cite{tremblay2020rain} propose a physics-based rain model to simulate photo-realistic rain at 8 different levels of severity. They also leverage the re-projected LiDAR-based depth measurements and apply a particle simulation framework~\cite{de2012fast} to approximate the real-world dynamics of raindrops. The degradation severity can be specified on the basis of rain intensity measured in mm/hr, ranging from light rain at $1\,\text{mm/hr}$ to heavy rain at $200\,\text{mm/hr}$. For a challenging scenario, we use the highest severity level at $200\,\text{mm/hr}$ in all our experiments to simulate heavy rain on KITTI data.

\minisection{Task 3 - Snow:} The final task we address in our sequence of weather conditions is snow. Hendrycks and Dietterich~\cite{hendrycks2019robustness} evaluate the robustness of various deep neural networks in several different scenarios. In their benchmark, they also introduce an approach to simulate snow on images which we use to superimpose snow on the KITTI images.

\subsection{Implementation Details}
Throughout all experiments we use the open source PyTorch implementation\footnote{\href{https://github.com/ultralytics/yolov3/tree/d353371}{https://github.com/ultralytics/yolov3}, commit: d353371} of the YOLOv3~\cite{redmon2018yolov3} object detector. For statistical correction of the batch normalization layers in YOLOv3 we use the official DUA implementation\footnote{\href{https://github.com/jmiemirza/DUA/tree/7a0240c}{https://github.com/jmiemirza/DUA}, commit: 7a0240c}. We use the default settings from the implementations except for the learning rate and batch sizes. For training all models we use a batch size of 16 while learning rates are handled by an early-stopping criteria. When the validation error does not improve for 5 epochs, we decrease the learning rate by a factor of 3. If the validation error still does not improve after 3 learning rate changes, we stop training the model further. The initial learning rate value is $0.01$.

\subsection{Experimental Protocols and Metrics}
For all experiments we report the mean average precision (mAP) over all object classes, evaluated at 50\% intersection over union (IOU). We follow this setting for reporting all object detection results unless stated otherwise.

In IL, it is common to provide metrics which evaluate over all the tasks seen so far while advancing through the sequence of tasks. Thus we adopt this protocol. In most IL setups, a weighted average over the task is provided, since the amount of classes and data might be variable. However, in our experimental setting we have the same amount of samples per weather condition (task), therefore a simple average over all seen tasks mAPs is sufficient.

Since our proposed domain-IL scenario is task-aware (\ie~we know the weather condition present in each image), we allow all compared methods to use this information and thus, apply or replace the corresponding task-specific parts of each model. 
For example, for Freezing we use the detector head trained for the specific weather condition. 
In Sec.~\ref{subsec:forward_backward_transfer}, we additionally analyse how well weather-specific statistics or models can perform across the different weather conditions.

\begin{table*}
    \centering
    \small
    \subfloat[Offline\label{tab:results:offline}]{
\begin{tabular}{c|c@{{ }$\rightarrow${ }}c@{{ }$\rightarrow${ }}c@{{ }$\rightarrow${ }}c}
  \toprule
   %Method & clear & fog & rain & snow \\
   \textbf{Method} & \textbf{clear} & \textbf{fog} & \textbf{rain} & \textbf{snow} \\
  \midrule
  Source-Only & {91.7} & 58.9 & 61.9 & 51.8 \\
  Freezing & {91.7} &61.4 &63.7 &54.0 \\
  Disjoint & {91.7} &56.7 &66.9 &71.4 \\
  Fine-tuning & {91.7} &59.4 &\textbf{80.0} &\textbf{76.2} \\
  \midrule
  DISC &{91.7} &\textbf{66.2} &68.8 &59.7 \\
  \midrule
  Joint-training &{91.7} &95.3 &97.2 &97.2 \\
  \bottomrule
  
  \end{tabular}}\hspace{1cm}
  \subfloat[Online\label{tab:results:online}]{
\begin{tabular}{c|c@{{ }$\rightarrow${ }}c@{{ }$\rightarrow${ }}c@{{ }$\rightarrow${ }}c}
  \toprule
  %Method & clear & fog & rain & snow \\
   \textbf{Method} & \textbf{clear} & \textbf{fog} & \textbf{rain} & \textbf{snow} \\
  \midrule
   Source-Only & 91.7$\pm$0.0 & 58.9$\pm$0.0 & 61.9$\pm$0.0 & 51.8$\pm$0.0 \\
  Freezing &91.7$\pm$0.0 &60.8$\pm$0.9 &63.6$\pm$0.3&53.6$\pm$0.4\\
  Disjoint & 91.7$\pm$0.0& 38.4$\pm$1.0&66.0$\pm$1.2&57.3$\pm$1.0\\
  Fine-tuning & 91.7$\pm$0.0&34.0$\pm$3.9&67.6$\pm$1.3&39.6$\pm$1.1\\
\midrule
  DISC & \textbf{91.7$\pm$0.0} & \textbf{66.2$\pm$0.0} & \textbf{68.8$\pm$0.0} & \textbf{59.7$\pm$0.0}\\
  \midrule
  Joint-training & 91.7$\pm$0.0&79.7$\pm$1.4&85.4$\pm$0.6&84.9$\pm$0.7\\
  \bottomrule
  
  \end{tabular}}
  \caption{Mean Average Precision (mAP@50) on KITTI averaged over all object classes. Following the common IL protocol, we report the results as the mean over the current and all previously seen tasks. (a) Offline setting, where all baselines are optimized until convergence. (a) Online setting, where all baselines are only trained for a single epoch. We report the mean and standard deviation over 10 runs.}
\label{tab:results}
\end{table*}

\subsection{Baselines}
\label{subsec:baselines_intro}
We propose to compare with several baselines which can be implemented in two distinct categories: offline and online. While offline approaches are allowed to train until convergence on each new task, online approaches are only trained for a single epoch on the new task. Thus, for offline approaches we can generally expect a better performance on the newest task, with an increased amount of forgetting of previous tasks. On the other hand, as online approaches are optimized only for a single epoch on the new tasks, it is expected that they maintain a stronger stability and reduced amount of forgetting. We evaluate all baselines in both online and offline settings. The baselines we consider are:
% OLD - The baselines we compare with can be divided into two distinct categories: offline and online. The main difference is that offline methods are trained until convergence on the new task while online baselines are only trained for a single epoch on the new task. For offline approaches we can generally expect that they will perform better for the new task they are optimized for, while a more severe forgetting of the previous tasks can be expected. On the other hand, as online baselines are optimized only for a single epoch on the new task, it is expected that they do not suffer significantly from forgetting. We evaluate all the baselines in both online and offline scenarios unless stated otherwise.

\minisection{Source-Only} maintains the initial model trained on clear weather data completely fixed throughout all tasks. This baseline does not perform any adaptation, modification or replacement to its parameters and serves as a lower bound.
% OLD - refers to the evaluations which we perform by keeping the initial model $\Phi\clear$ completely fixed after training on Task-0 (\ie the clear weather condition) and evaluating on all other tasks. This baseline does not perform any adaptation to changing weather conditions at all and serves as the lower bound for verification.

\minisection{Freezing} only allows to update the parameters for the three YOLOv3 heads -- which are specialized on detecting objects at different spatial resolutions -- while keeping the representations learned by the remaining layers frozen. To train the separate sets of heads for each of the tasks, we use the learning rate at which the training for the initial task terminated. Similar to DISC, freezing is also a zero-forgetting approach.
% OLD - refers to the evaluations which we perform by only training the three YOLOv3 heads which are specialized for detecting objects at different spatial resolutions, while keeping the representations learned by the encoder for Task-0 frozen. For training the heads we use the learning rate at which the training for Task-0 terminates. Similar to DISC, freezing is also another zero-forgetting approach.

\minisection{Disjoint} trains a separate model for each task. During inference, the model which matches the corresponding task-ID is used. To train this baseline, we initialize each of the disjoint YOLOv3 models with weights pre-trained on the MS COCO~\cite{lin2014microsoft} object detection dataset and fine-tune for the corresponding weather condition until convergence.
% OLDer - refers to the evaluations which we perform by training individual models for each task separately. Then, these models are evaluated on all other tasks. For this, we always start with a YOLOv3 model pre-trained on the MS COCO~\cite{lin2014microsoft} object detection dataset and finetune it for the corresponding weather condition until convergence.

\minisection{Fine-tuning} learns each task incrementally in a fully-supervised manner.
%Once a task has been learned until convergence, the next task will be initialized from the previous task parameters and learn the corresponding new task. The learning rate used is the one at which the initial task terminated.
When a new task arrives, the current model will be fine-tuned on the task-specific training data until convergence. The learning rate used is the one at which the initial task terminated.
% OLD - refers to the evaluations we perform by learning new tasks incrementally in a fully supervised manner until convergence. For example, the four tasks are learned in a sequential manner by using the model which has the weights finetuned for the previous task. We use the learning rate at which the training for the Task-0 terminates for finetuning the model to a new task.

\minisection{Joint-training} trains the model with all data from all tasks seen so far in a fully-supervised manner, breaking one of the key incremental learning properties and thus, provides an upper bound. The objective is to provide a comparison to a model which can learn a weather-invariant representation which performs well under all weather conditions.

\section{Results}
We now present detailed results comparing DISC to different baselines, for both offline and online training.
%introduced in Section~\ref{subsec:baselines_intro}.

\subsection{Comparison in the Four-Task Scenario}
First, we evaluate the four different weather conditions (introduced in Sec.~\ref{subsec:dataset-tasks}) when presented in the domain-incremental learning setting (described in Sec.~\ref{sec:domain-il}). The results are summarized in Table~\ref{tab:results}.
%In Table~\ref{tab:results} we provide a comparison between the baselines and DISC for both offline and online training.

In Table~\ref{tab:results:offline} we compare the \emph{offline setting}, where all baselines are trained in a supervised manner until convergence. Results show the mAP@50 averaged over all object classes, with DISC obtaining the best results after learning the fog task, and Fine-tuning obtaining the best results after the remaining tasks of the sequence. DISC obtains better results than the other zero-forgetting approach, Freezing, across all the sequence, showing that adapting the intermediate layer representations is better than freezing them. 
%The reduced performance after the rain and snow tasks in comparison to the Disjoint and Fine-Tuning baselines is expected, since DISC does not require of any of the computationally expensive training. Finally, the results from Joint-Training being quite higher than any of the other compared methods opens the door to further improved strategies to come.
Due to our challenging adverse weather simulation settings, there is a notable gap between all approaches and the upper performance bound (\ie~Joint-training), which opens the door for further improved strategies in the future.
Note that the performance gap in comparison to the Disjoint and Fine-tuning baselines after the rain and snow tasks is expected, because in contrast to DISC, both baselines can fully adapt the detection model. DISC, however, provides a simple and efficient way to notably improve the initial model performance in varying weather conditions, without requiring extensive re-training or memory banks.

% OLD - In Table~\ref{tab:results} we provide results for DISC and comparison with baselines in the four tasks scenario. In Table~\ref{tab:results:offline} we compare DISC to baselines when we train them in an offline manner. In this scenario all the baselines are trained in a supervised manner until convergence.

In Table~\ref{tab:results:online}, we provide results for the \emph{online setting}, where all baselines are trained for a single epoch on subsequent tasks. To ensure a fair comparison in this setting, all models are initialized with a model pre-trained on the clear weather condition.
In this case, DISC obtains the best results across all tasks and throughout the whole sequence. This demonstrates that DISC is well suited for the practical online scenario, since it only needs to store the statistics after forwarding the samples through the network, and does not need any other training for subsequent tasks.
%without any other training involved after the first task is learned.
Comparing all methods to the lower bound (\ie~Source-Only) shows that this learning setting is not trivial: Without using exemplars or any other technique to avoid catastrophic forgetting, both Fine-tuning and Disjoint degrade significantly at the most challenging fog and snow conditions.
%Freezing obtains the second best results for both fog and snow, since freezing the representations allows for a more stable training under this setting.
% OLD - In Table~\ref{tab:results:online}, we provide results for DISC when compared with baselines trained in an online manner. Here all the baselines are also trained in a supervised manner but for a single epoch.
% OLD - Results show that despite of its simplicity and not requiring any labels from task other than `clear', DISC is able to be competitive in the offline scenario while outperforming other expensive baselines in an online scenario. We also want to point out that as DISC is an online incremental learning approach thus its comparison with offline baselines is more fair.

\begin{figure*}
\vspace{3pt}
    \centering
    \includegraphics[scale=0.58, trim=45 4 45 0, clip]{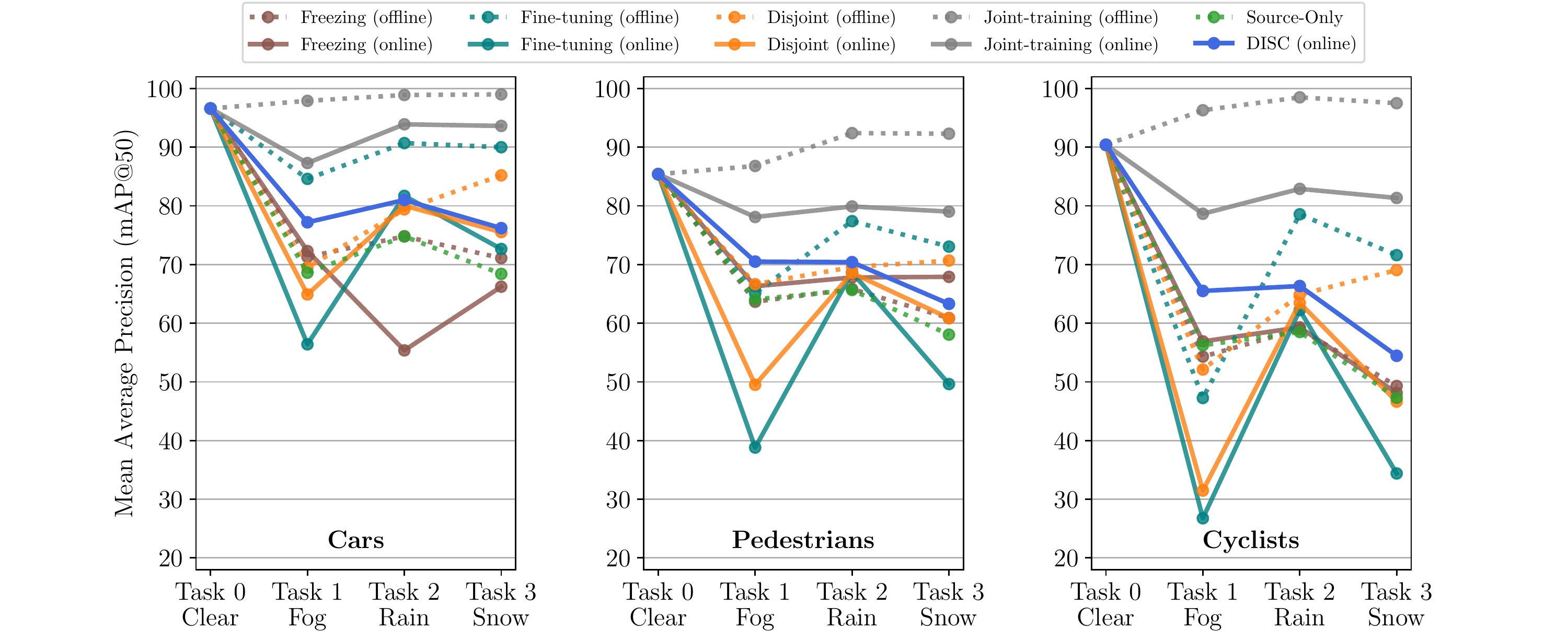}
    \caption{Mean Average Precision (mAP@50) for the three most common traffic participant classes, \ie~cars, pedestrians, and cyclists. We show the results for all baselines for both online and offline training.
    %The results for all the baselines are reported while evaluating them in an offline manner. 
    We follow the common IL evaluation protocol and report the performance at a particular task as the mean over this task and all previously seen tasks.
    %We again follow the common evaluation protocol of IL where mean is reported of the results obtained at a particular task and all the previously seen tasks.
    }
    \label{fig:results_3_classes}
\end{figure*}

\subsection{Interaction between Tasks}
\label{subsec:forward_backward_transfer}
To investigate how different the weather condition tasks are, we conduct an experiment where we use the task-specific statistics/models on all other tasks.
%To provide additional insights on how different the weather condition tasks are, we conduct an experiment where we use the task-specific parts of the model on all other tasks.
For DISC, we use each weather-specific statistics and evaluate the performance under all other weather conditions. Similarly for Disjoint, we evaluate each of the separate weather models on all weather tasks. Through this experiment we want to observe how good the learned representations are when presented with a significant domain shift, which could happen by incorrectly estimating the current weather task-ID.
%To get some insight on how different are the weather condition tasks, we conduct an experiment where we evaluate them while using the task-specific parts of the model from each of them. For DISC, we use each of the statistics for each weather and evaluate the performance of those tas-specific models under all weather conditions. For Disjoint, we evaluate each of the separate weather models on the data of each of the 4 tasks. With that, we want to observe how good are the learned representations when presented with a domain shift from the task they initially learned.

\begin{figure}
    \centering
    \subfloat[\centering DISC \label{fig:confusion_matrix_disc}]{{\includegraphics[scale=0.45, trim=15 14 14 14, clip]{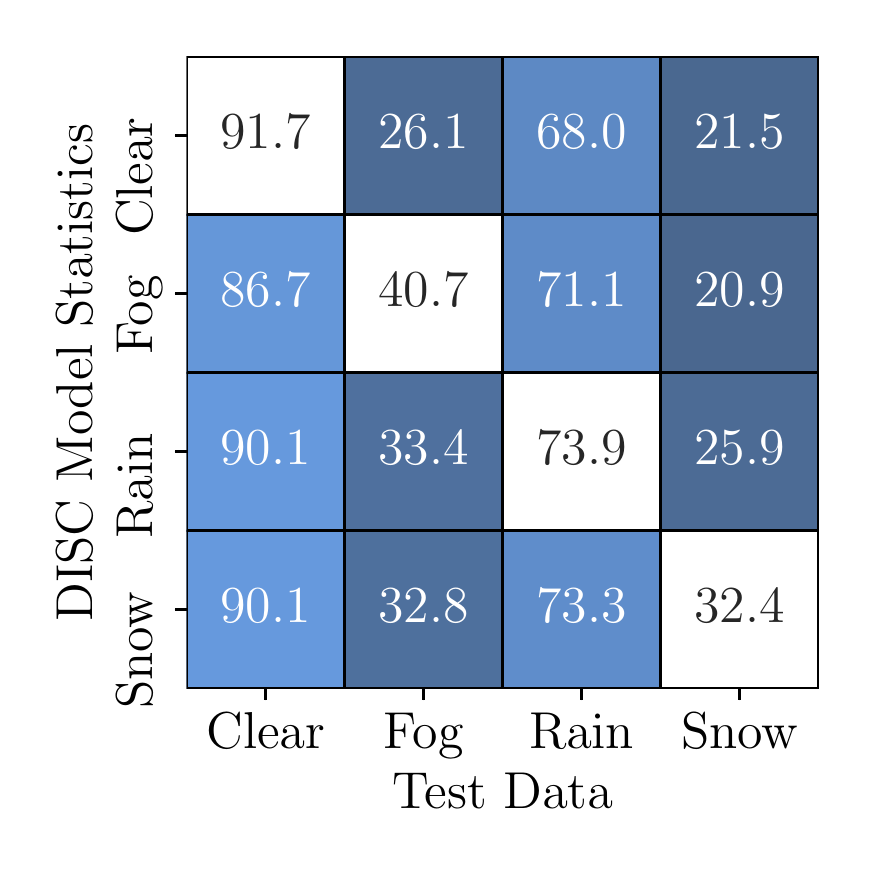} }}%
    \hfill
    \subfloat[\centering Disjoint\label{fig:confusion_matrix_disjoint}]{{\includegraphics[scale=0.45, trim=15 14 14 14, clip]{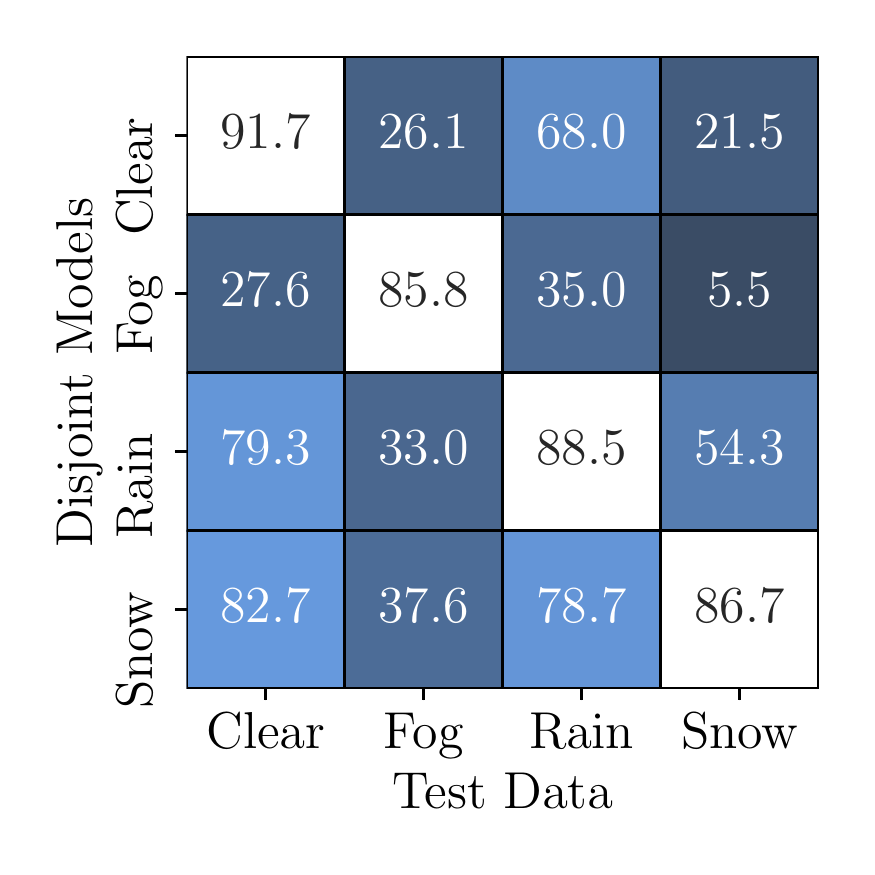} }}%
    \caption{Cross-task evaluation using mAP@50 averaged over all classes. We apply (a)~DISC's weather-specific statistics and (b)~the per-weather optimized models on all different weather conditions. Entries along the diagonal show results at the condition for which the corresponding statistics/models were trained for.}
    %\caption{Mean Average Precision (mAP@50) averaged for all classes in the KITTI dataset. (a) DISC evaluations performed while equipping the models with specific weather statistics and evaluated for all weather conditions. (b) Disjoint models optimized for specific weather and evaluated on all weather conditions in our scenarios.}%
    \label{fig:confusion_matrix}%
\end{figure}

The results of this experiment are presented in Fig.~\ref{fig:confusion_matrix}.
The diagonal entries show the results obtained when using the methods with the correct task-ID, while the first row contains the results corresponding to the Source-Only baseline. In all cases, using the statistics/models for the weather condition they are trained on provides the best results. For both DISC and Disjoint, we can observe that fog and snow are the most challenging weather conditions, most likely due to some objects becoming barely visible in the deep fog/heavy snow. The Disjoint results demonstrate that, obviously, a weather-specific model has a clear advantage on highly adverse conditions (especially snow). This can be attributed to the large visual appearance gap (\eg~consider clear weather \emph{vs.} almost "white-out" snow scenario) which seemingly requires different feature representations to be learned by the detection model.
%OLD - The results of this experiment are presented in Fig.~\ref{fig:confusion_matrix}. The tables represent the combination of the task-specific part of the model being used to evaluate on each of the weather conditions. The diagonal contains the results obtained when using the methods with the correct task-ID, while the first row contains the results corresponding to the Source-Only baseline. In both cases, we can observe that the fog weather is the most challenging one, probably due to some of the objects becoming barely visible under the deep fog. In the case of snow, we can see that DISC has some trouble adapting to it, while Disjoint shows that a snow-specific model can achieve very good results. This could be due to snow requiring different representations than the ones used for clear. In all cases, using the task-specific part from the weather condition of the data being evaluated provides the best results.

Considering the DISC results in this cross-task experiment, we see that the statistics learned from rain and snow conditions seem to have good transferability, as the performance is always comparable or better than using the initial clear weather statistics (\ie~Source-Only results). Fog statistics, on the other hand, perform significantly better than clear weather statistics only when evaluated on fog data. For both DISC and Disjoint, we see that using fog-specific statistics/model on snow data, the performance decreases. This degradation is especially significant for Disjoint. In general, it seems that Disjoint models are very good at solving the weather condition they have been trained on, while mostly having less transferability to other domains than DISC.
%OLD - For DISC, using statistics from rain or snow, is always comparable or better than using the statistics from the initial task with clear. Therefore, rain and snow seem to have more transferability, while fog statistics only do significantly better when evaluated on its corresponding fog data. It is worth noticing that when using fog-specific statistics/model on snow data, performance decreases, specially for Disjoint. In general, it seems that Disjoint models are very good at solving the weather condition they have been trained on, while mostly having less transferability to other domains.

\subsection{Results for Individual Classes}
Following the common evaluation protocol for object detection in autonomous driving scenarios, \eg~\cite{ku2018joint, lang2019pointpillars,mirza2021norm}, we separately analyse the results for the three most frequently occurring traffic participants, \ie~cars, pedestrians, and cyclists.
From Fig.~\ref{fig:results_3_classes} we can observe several trends: In general, the performance for all classes drops after adjusting to fog. Moving to the next weather change (rain), we see that the performance of the zero-forgetting methods (DISC \& Freezing) stays more or less the same, whereas offline approaches improve (as they can adjust the model until convergence) and competing online approaches degrade even further. Finally, when encountering the challenging snow scenario all approaches -- except for the offline Disjoint variant -- drop in performance.
Overall, these results also confirm that a realistic and practical online learning scenario is significantly more challenging than the offline setting.
%
%The general trend for all classes is that after learning task~1 (fog), the performance drops. After moving further on to Task-2 (rain), the zero-forgetting approaches stay more or less the same, but the ones that train all the model increase their performance. Finally, all approaches except for Disjoint (offline) drop in performance when learning Task-3 (snow).

While this behaviour is similar to the previously observed results in Table~\ref{tab:results:offline} and~\ref{tab:results:online}, here we can observe some object class-specific differences.
When driving in deep fog, reliably detecting cyclists is most challenging for all approaches, indicated by the large performance drops for this class.
%When driving in deep fog, it is more complicated for all approaches to detect cyclists than other classes, indicated by the large performance drop for this class. 
Furthermore, considering the upper performance bound (\ie~Joint-training), future research should focus on pedestrian detection in adverse weather conditions to enable robust and safe autonomous driving systems.
%we observe that the pedestrian class is more challenging than car or cyclist.

% OLD - Results for DISC and other baselines in the offline scenario are provided in Figure~\ref{fig:results_3_classes}. We again provide an average of results obtained for each class over all the tasks seen so far at each stage. The behaviour of all the baselines and DISC for all the tasks is mostly consistent. For most of the results we can see a sharp drop in performance when we move to Task-1 (fog). 
% This behaviour can also be correlated with natural scenes when dense fog is encountered. Even human drivers might find it difficult to drive in dense foggy conditions. 

% \begin{figure*}
%     \centering
%     \subfloat[\centering Cars \label{subfig:individual_cars}]{{\includegraphics[scale=0.45, trim=2 2 2 3, clip]{figs/car.pdf} }}%
%     \hfill
%     \subfloat[\centering Pedestrians \label{subfig:individual_pedestrians}]{{\includegraphics[scale=0.45, trim=2 2 2 3, clip]{figs/pedestrian.pdf} }}%
%     \hfill
%     \subfloat[\centering Cyclists\label{subfig:individual_cyclist}]{{\includegraphics[scale=0.45, trim=2 2 2 3, clip]{figs/cyclist.pdf} }}%
%     \caption{Results for three common classes in the KITTI dataset.}%
%     \label{fig:results_3_classes}%
% \end{figure*}

\section{Conclusion}
We proposed a novel zero-forgetting domain-incremental learning approach to efficiently address varying weather conditions in autonomous driving scenarios.
By only correcting the weather-specific domain statistics when encountering weather changes, we achieve robust object detection even in adverse weather conditions. Our evaluations on challenging consecutive weather changes show promising results of our DISC approach for both online and offline learning scenarios.

\paragraph{Acknowledgments}
\label{sec:acknowledgement}
This work was partially funded by the Christian Doppler Laboratory for Embedded Machine Learning and the Austrian Research Promotion Agency~(FFG) under the project High-Scene~(884306). Marc Masana acknowledges the support by the ``University SAL Labs'' initiative of Silicon Austria Labs (SAL).
% We gratefully acknowledge the financial support by the Austrian Federal Ministry for Digital and Economic Affairs, the National Foundation for Research, Technology and Development and the Christian Doppler Research Association. This work was also partially funded by the Austrian Research Promotion Agency~(FFG) under the project High-Scene~(884306).
%In this paper we have proposed a novel zero-forgetting Online Domain Incremental learning approach DISC which only stores the first and second order statistics for the different weather conditions at the time of training. 
%At the time of testing, given the Task-ID, DISC only `plugs in' the domain statistics for the particular task (weather condition) while keeping all the other model parameters fixed.
%We test DISC for the task of object detection in a challenging domain incremental learning scenario where each weather condition is considered as a  different task. By only plugging in the relevant statistics for different weather scenarios, DISC shows promising results when compared to baselines both in offline and online scenarios. 

%%%%%%%%% REFERENCES
{\small
\bibliographystyle{ieee_fullname}
\bibliography{egbib}
}

\end{document}